%% file: main.tex
\documentclass{article}

\usepackage[preprint]{neurips_2026}

\usepackage{booktabs}
\usepackage[utf8]{inputenc} %
\usepackage{amsthm}
\usepackage[T1]{fontenc}    %
\usepackage{hyperref}       %
\hypersetup{hidelinks}
\usepackage{url}            %
\usepackage{booktabs}       %
\usepackage{amsfonts}       %
\usepackage{nicefrac}       %
\usepackage{microtype}      %
\usepackage{xcolor}         %
\usepackage{amsmath}
\usepackage{pifont}
\usepackage[table]{xcolor}
\usepackage{ulem}
\usepackage{array}
\usepackage{tabularx}
\usepackage{graphicx}
\usepackage{multirow}
\usepackage{caption}

\usepackage{ulem}
\usepackage{booktabs}      %
\usepackage{xcolor}        %
\usepackage{colortbl}      %
\usepackage[table, dvipsnames]{xcolor}
\usepackage{tabularx}

\usepackage{nicematrix} %
\newcommand{\best}[1]{%
    \cellcolor{green!20}\textbf{#1}%
}

\newcommand{\second}[1]{%
    \cellcolor{blue!20}%
    \uline{#1}%
}
\newcolumntype{Y}{>{\centering\arraybackslash}X} %

\newif\ifshowrevisions
\showrevisionsfalse
\ifshowrevisions
\definecolor{revisionblue}{RGB}{0,70,180}
\else
\definecolor{revisionblue}{gray}{0}
\fi
\newcommand{\rev}[1]{{\color{revisionblue} #1}}
\newenvironment{revision}{\begingroup\color{revisionblue}}{\endgroup}
\usepackage{needspace}
\newcommand{\papertablestyle}{%
  \small
  \setlength{\tabcolsep}{4pt}%
  \renewcommand{\arraystretch}{1.05}%
  \setlength{\aboverulesep}{0.25ex}%
  \setlength{\belowrulesep}{0.40ex}%
}

\title{MIRA: Reinforcing Multimodal Reasoning via Deceptive Contextual Augmentation}

\author{%
  \rev{\textbf{Zhihan Yin}$^{1,2}$\quad
  \textbf{Jianxin Liang}$^{1}$\quad
  \textbf{Yifeng Yao}$^{1}$\quad
  \textbf{Nonghai Zhang}$^{1,2}$}\\
  \rev{\textbf{Bingyue Peng}$^{2}$\quad
  \textbf{Chunguang Qie}$^{2}$\quad
  \textbf{Huishuai Zhang}$^{1,3,*}$\quad
  \textbf{Dongyan Zhao}$^{1,3,4,*}$}\\
  {\normalfont\rev{$^{1}$Peking University\quad $^{2}$ByteDance\quad
  $^{3}$Wangxuan Institute of Computer Technology}}\\
  {\normalfont\rev{$^{4}$National Engineering Research Center of New Electronic Publishing Technologies}}\\
  {\normalfont\rev{$^{*}$Corresponding authors.}}
}

\begin{document}

\maketitle

\begin{abstract}
While Reinforcement Learning with Verifiable Rewards (RLVR) has significantly advanced multimodal reasoning, existing frameworks suffer from a critical “perception–reasoning gap.” Due to an over-reliance on seemingly relevant visual tokens, even minor perceptual perturbations can propagate through the reasoning process, leading to compounding hallucinations. To address this issue, we propose MIRA, a training framework that improves robustness to erroneous visual contexts while enhancing logical reasoning ability. By injecting filtered, deceptive visual contexts during training, we construct challenging “hard traps” that stress-test the model’s robustness to misleading information. Furthermore, we introduce a group-wise reflection-triggering mechanism that enables the model to autonomously detect and rectify inconsistencies between visual cues and logical constraints, which is ultimately internalized as an intrinsic policy behavior. Importantly, MIRA does not rely on human annotations. Experimental results across multiple reasoning and perception benchmarks show that MIRA yields an average 7\% improvement on in-domain benchmarks, with consistent improvements on vision-centric benchmarks. These findings highlight that active error recovery is essential for robust multimodal reasoning. Comprehensive ablation studies and analyses further provide insights into why MIRA is effective. Code is available at \href{https://github.com/Hans-M-Yin/mira-neurips26}{Github}.
\end{abstract}

\input{sections/intro}

\input{sections/related_works}

\input{sections/preliminaries}

\input{sections/method}

\input{sections/experiment}

\input{sections/analysis}

\section{Conclusion}
In this paper, we introduced MIRA, a novel training framework designed to bridge the "perception-reasoning gap" in multimodal large language models. By systematically constructing deceptive visual contexts and employing an intra-group reflection-triggering mechanism, we enable models to autonomously detect and rectify perceptual inconsistencies without relying on human-annotated chain-of-thought data. Experimental results across diverse benchmarks demonstrate that MIRA significantly enhances both logical reasoning and visual robustness, outperforming several recent reinforcement learning baselines. Furthermore, our analysis confirms that this self-correction capability is internalized as an intrinsic policy behavior, allowing the model to re-attend to visual evidence even in the absence of explicit triggers. While the current validation of MIRA is focused on a 7B-parameter scale due to computational constraints, we aim to explore the scaling properties of this self-correction mechanism on larger foundation models in future work. Ultimately, these findings underscore the necessity of active error recovery for reliable multimodal reasoning and provide a scalable path toward more resilient vision-language systems.

\clearpage
\begin{ack}
The work was supported in part by Beijing Natural Science Foundation (L2608055) and National Natural Science Foundation of China (Grant No.~62576015). The authors sincerely thank the anonymous reviewers for offering invaluable suggestions.
\end{ack}

\bibliographystyle{plainnat}
\bibliography{reference}

\clearpage
\input{sections/appendix}

\end{document}

%% file: sections/intro.tex
\section{Introduction}
Recent advances in large language models (LLMs) have demonstrated that Reinforcement Learning with Verifiable Rewards (RLVR) can substantially enhance  reasoning capabilities, enabling strong performance on complex tasks such as mathematics and programming~\citep{shao2024deepseekmathpushinglimitsmathematical,Guo_2025}. As a natural extension of LLMs, Vision-Language Models (VLMs) are required to tackle more challenging multimodal reasoning scenarios~\citep{liu2023visualinstructiontuning,bai2025qwen3vltechnicalreport}, where models must not only perform long-context reasoning in the language space but also effectively integrate and interpret information across modalities.

To address this challenge, a series of multimodal reasoning approaches have been proposed, achieving notable progress. Methods such as Vision-R1~\citep{huang2026visionr} and VL-R1~\citep{zhang2025r1vllearningreasonmultimodal} successfully transfer the long-chain reasoning capabilities of LLMs to the multimodal domain through RLVR-based post-training, enabling VLMs to exhibit sophisticated reasoning behaviors~\citep{meng2025mmeurekaexploringfrontiersmultimodal,wangvl}. However, while these methods strengthen logical reasoning, they largely overlook the model’s ability to reflect on and correct errors in the visual space~\citep{yang2025lookbackimplicitvisualrefocusing}. Prior studies show that approximately 70\% of errors in multimodal reasoning originate from visual perception~\citep{wang2026perceptionaware,rezaei2026cppocontrastiveperceptionvision}. Once a model misinterprets visual information, it reasons based on incorrect premises, causing errors to propagate and compound throughout the reasoning process, ultimately leading to consistently incorrect conclusions. This further highlights that accurate visual understanding is a fundamental prerequisite for reliable multimodal reasoning.

Existing efforts attempt to mitigate this issue by enhancing visual grounding~\citep{yu2026perceptionr}, reweighting visual tokens~\citep{lu2026bridging,wang2026visuallyguidedpolicyoptimizationmultimodal} and etc. While these approaches improve visual understanding to some extent, they do not explicitly equip models with the ability to perform self-correction in the visual space, and thus fail to fundamentally prevent error propagation. Moreover, these methods often require substantial modifications to the RLVR training pipeline, raising concerns about their scalability and generalization to broader settings.

To address these limitations, we propose \textbf{MIRA} (MIsleading-context Robustness Augmentation), a multimodal reasoning framework centered on constructing deceptive contextual augmentations. Rather than directly improving the model’s visual perception accuracy, our goal is to equip the model with the ability to recover from initial misinterpretations, enabling it to revise incorrect perceptions and still arrive at correct conclusions. Specifically, we simulate potential perception errors by injecting perturbed visual descriptions into the context, and introduce a group-wise enforced reflection mechanism that encourages the model to correct such errors during reasoning. Notably, our method enhances visual self-correction capabilities without requiring additional supervised fine-tuning (SFT) cold-start data.

Technically, MIRA first samples \rev{question-relevant visual descriptions} using the policy model, and then employs an expert model to generate adversarial perturbations. A filtering step is applied to remove perturbations that do not affect the final answer, ensuring the effectiveness of the training signal. These perturbed descriptions are then injected into the context as if they were the model’s own previous outputs, thereby simulating realistic error scenarios. To prevent the model from ignoring contextual inconsistencies and to explicitly encourage corrective behavior, we introduce reflection triggers within sample groups, allowing the model to internalize the benefits of visual self-correction as part of its policy. Importantly, our method applies specialized training only to augmented data and does not require additional human annotations. By retaining a proportion of original data during training, we preserve the model’s generalization ability and mitigate potential overfitting or policy degeneration.

Extensive experiments demonstrate the effectiveness of MIRA. Our method improves the base model by \rev{approximately 7 percentage points on average} across multiple mathematical benchmarks. Moreover, without introducing additional general-purpose visual data, MIRA achieves consistent gains on several multimodal benchmarks and outperforms a range of existing approaches. We further conduct comprehensive ablation studies to analyze the contributions of each component.

%% file: sections/related_works.tex
\section{Related Works}
\textbf{Vision-Language Models} Vision-Language Models (VLMs) have made significant progress in aligning visual and textual representations through large-scale pretraining and modular architectures~\citep{liu2023visualinstructiontuning,bai2025qwen25vltechnicalreport,bai2025qwen3vltechnicalreport}. Existing approaches generally follow two paradigms: (1) end-to-end training on paired multimodal data, jointly optimizing all parameters~\citep{wang2024emu3nexttokenpredictionneed}, and (2) freezing the language backbone while introducing lightweight modules (e.g., adapters or projection layers) to bridge modalities~\citep{liu2023visualinstructiontuning,zhu2024minigpt,li2023blip2bootstrappinglanguageimagepretraining,alayrac2022flamingovisuallanguagemodel}. More recent works explore unified multimodal architectures to improve generality and scalability~\citep{deng2025emergingpropertiesunifiedmultimodal,xin2025luminadimooomnidiffusionlarge,li2025uniworldv2reinforceimageediting}. Despite these advances, most methods primarily focus on cross-modal alignment and fusion, while overlooking the modeling of error propagation in visual perception. As a result, VLMs remain sensitive to initial perceptual errors, which can cascade through subsequent reasoning.

\textbf{Multimodal Reinforcement Learning} Reinforcement learning has emerged as an effective paradigm for enhancing reasoning abilities in large models without fine-grained supervision~\citep{shao2024deepseekmathpushinglimitsmathematical,Guo_2025}. RLHF aligns models with human preferences~\citep{ouyang2022traininglanguagemodelsfollow}, while outcome-based methods such as GRPO demonstrate that verifiable rewards alone can drive strong long-chain reasoning~\citep{shao2024deepseekmathpushinglimitsmathematical,yu2026dapo,zheng2025groupsequencepolicyoptimization}. These ideas have been extended to multimodal settings via reinforcement learning with verifiable rewards (RLVR), significantly improving reasoning capabilities~\citep{huang2026visionr,shen2025vlmr1stablegeneralizabler1style,zhang2025r1vllearningreasonmultimodal,xia2025visionaryr1mitigatingshortcutsvisual,li2026visionsr,leng2025mmr1,fan2026sophiavlr}. However, existing methods exhibit a clear limitation. Approaches focused on reasoning (e.g., Vision-R1~\citep{huang2026visionr}) often neglect perception robustness, leading to visual hallucinations, while perception-enhanced RL methods~\citep{wang2026visuallyguidedpolicyoptimizationmultimodal,ouyang2025conanprogressivelearningreason,pan2026through,xiao2026perceptionr,liu2026noisyrollout,zhu2026shuffler} improve visual perception but is still unable to deal with  existing perceptual errors propagate through the reasoning process. Consequently, a trade-off emerges between perception and reasoning. In contrast, our method jointly improves both, while enabling visual-space self-reflection and error correction without supervised fine-tuning.

%% file: sections/preliminaries.tex
\section{Preliminaries}

Let $\pi_{\theta}$ denote a policy VLM parameterized by $\theta$. Given a multimodal input consisting of an image $I$ and a textual query $Q$, the model generates an output sequence $y = (y_1, \dots, y_N)$. The conditional probability of generating $y$ is defined as:
\begin{equation}
\pi_{\theta}(y \mid I, Q) = \prod_{t=1}^{N} \pi_{\theta}(y_t \mid I, Q, y_{<t}),
\end{equation}
where $\pi_{\theta}(y_t \mid I, Q, y_{<t})$ denotes the probability of generating the $t$-th token conditioned on the input and previous tokens. In reinforcement learning, the model receives a scalar reward $r(y, I, Q)$ based on the generated output $y$. The objective is to maximize the expected reward:
\begin{equation}
\theta^{*} = \arg\max_{\theta} \ \mathbb{E}_{(I, Q) \sim \mathcal{D}} \left[ \mathbb{E}_{y \sim \pi_{\theta}(\cdot \mid I, Q)} \left[ r(y, I, Q) \right] \right].
\end{equation}
\textbf{Group Relative Policy Optimization (GRPO)} samples a group of rollouts $y_i \sim \pi_{\theta_{\mathrm{old}}}(\cdot \mid I, Q)$ from the current policy, and refines the objective by normalizing rewards within each group. Specifically, rewards are standardized at the group level to compute relative advantages, which are then used to guide policy updates. Finally, the GRPO objective is defined as:
\begin{equation}
\label{eq:grpo}
\begin{aligned}
\mathcal{J}_{\mathrm{GRPO}}(\theta)
= &\ \mathbb{E}_{(I, Q) \sim p_{\mathcal{D}}, \ \{y_i\}_{i=1}^n \sim \pi_{\theta_{\mathrm{old}}}(\cdot \mid I, Q)}
\Bigg[ \\
&\ \frac{1}{n} \sum_{i=1}^{n}
\min \Bigg(
\frac{\pi_{\theta}(y_i \mid I, Q)}{\pi_{\theta_{\mathrm{old}}}(y_i \mid I, Q)} \hat{A}_i, \quad \mathrm{clip}\Bigg(
\frac{\pi_{\theta}(y_i \mid I, Q)}{\pi_{\theta_{\mathrm{old}}}(y_i \mid I, Q)},
1 - \epsilon,\ 
1 + \epsilon
\Bigg) \hat{A}_i
\Bigg)
\Bigg]
\end{aligned}
\end{equation}

where $\epsilon$ is the clipping range. The Kullback-Leibler divergence term between $\pi_{\theta}$ and $\pi_{\mathrm{old}}$ that penalizes large deviations from a reference policy $\pi_{\mathrm{ref}}$ is omitted for simplicity in Equation~\ref{eq:grpo}.

%% file: sections/method.tex
\section{Method}

\begin{figure}[t]
\centering
\includegraphics[width=0.99\linewidth]{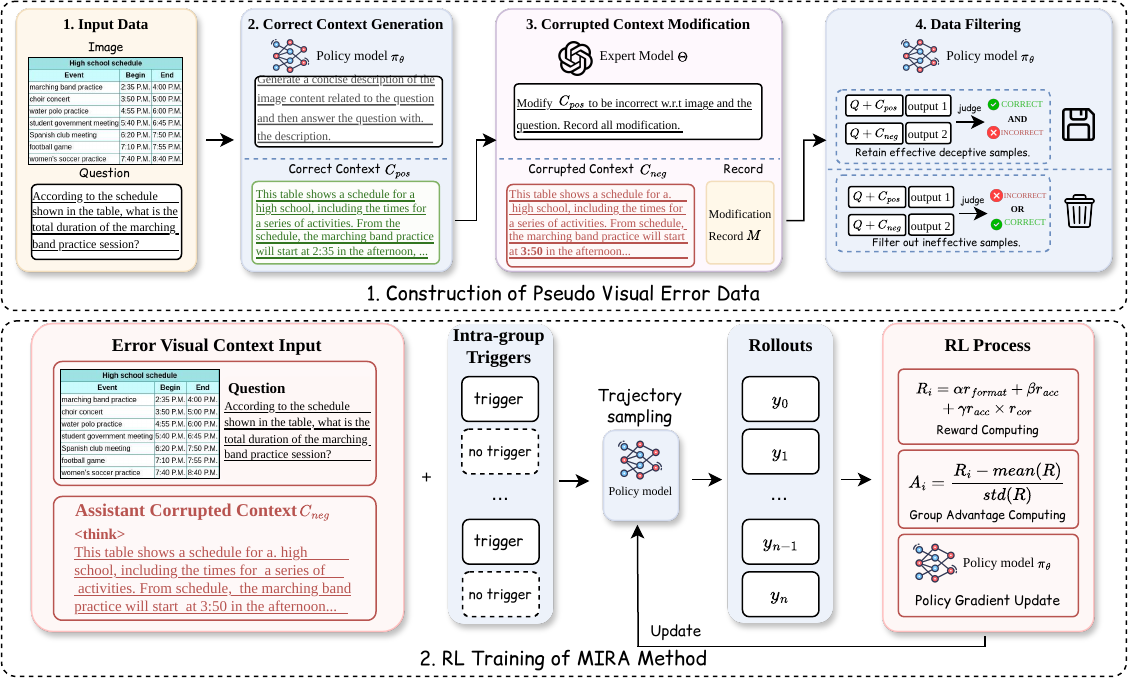}
\caption{Overview of MIRA. MIRA consists of two parts. The top illustrates the construction of deceptive visual context using the policy model and an expert model. The bottom shows the training process; compared to GRPO, the key changes are modified inputs and the use of intra-group triggers.}
\label{fig:pipeline}
\end{figure}

Recent studies suggest that, in complex multimodal reasoning tasks, errors often stem from imperfect visual understanding rather than purely logical failures~\citep{wang2026perceptionaware,rezaei2026cppocontrastiveperceptionvision}. Such inaccuracies in visual perception can propagate and negatively affect downstream reasoning.

Instead of enhancing visual perception by improving the vision encoder, our approach focuses on strengthening the interaction between visual and linguistic representations. Specifically, while reinforcement learning improves the model's reasoning ability, we introduce a visual-space reflection mechanism that enables the model to revisit and refine its visual understanding during inference. This allows the model to remain stable in disturbed visual context, leading to more robust reasoning.

\subsection{Construction of Deceptive Visual Context}
\label{sec:data_construction}
We construct deceptive visual-context data to encourage the model to re-examine its visual understanding during generation. \textcolor{revisionblue}{Specifically, we prepend to the model input a description of the image content disguised as a model-generated output. This description is intended to mislead the model's normal visual information extraction and  reasoning process.}

Given an input pair $(I, Q)$ \textcolor{revisionblue}{sampled from the publicly available dataset $\mathcal{D}_0$}, we first prompt the policy model $\pi_\theta$ to generate a visual description summarizing the visual information relevant to the question:
\begin{equation}
C_{\mathrm{pos}} \sim \pi_\theta(\cdot \mid I, Q, q_{\mathrm{des}}),
\end{equation}
where $q_{\mathrm{des}}$ denotes the corresponding system prompt. We intentionally use the same policy model, rather than a stronger external model, to reduce distribution mismatch and ensure that the generated description is naturally utilized during subsequent reasoning. \rev{$C_{\mathrm{pos}}$ is not guaranteed to be fully correct; the correction reward supervises only the expert-injected errors in $M$.} 

Next, an expert model $\Theta$ modifies $C_{\mathrm{pos}}$ into a plausible but incorrect description $C_{\mathrm{neg}}$, conditioned on the image, question, and ground-truth:
\begin{equation}
(C_{\mathrm{neg}}, M) = \Theta(q_{\mathrm{modify}}, I, Q, C_{\mathrm{pos}}),
\end{equation}
where $M$ records the applied modifications as pairs of before-and-after descriptions. This $C_{\mathrm{neg}}$ is consistent with $C_{\mathrm{pos}}$ in both structure and overall content, but differs in specific visual description details, and such differences affect the final answer to the question. Figure~\ref{fig:pipeline} illustrates a example.

To ensure the injected errors effectively influence the model, we retain only samples where \rev{a text-only verifier answers correctly from $C_{\mathrm{pos}}$ and incorrectly from $C_{\mathrm{neg}}$, without access to the image}. The resulting dataset is:
\begin{equation}
\mathcal{D}_{\mathrm{aug}} =
\left\{
(I, Q, C_{\mathrm{neg}}, M)
\ \middle| \
\rev{\hat{y}_{\mathrm{pos}}=y^*,\ \hat{y}_{\mathrm{neg}}\ne y^*}
\right\},
\end{equation}
where $\hat{y}_{\mathrm{neg}}$ and $\hat{y}_{\mathrm{pos}}$ denote the \rev{verifier predictions, compared with ground truth $y^*$,} conditioned on $C_{\mathrm{neg}}$ and $C_{\mathrm{pos}}$, respectively. \rev{We retain all original samples in $\mathcal{D}_0$ and append their augmented versions in $\mathcal{D}_{\mathrm{aug}}$, without introducing new questions:}
\begin{equation}
\mathcal{D_{\mathrm{MIRA}}} = \mathcal{D}_{\mathrm{aug}} \cup \mathcal{D}_{0}
\end{equation}

\subsection{Intra-group Refinement Trigger}

When deceptive visual context is introduced, the policy model, which has not been explicitly trained on such input patterns, may exhibit suboptimal behaviors. It may either over-rely on the context for reasoning or ignore it entirely by implicitly reconstructing a plausible caption from visual features, which weakens the intended perturbation effect.

Inspired by VL-Rethinker~\citep{wangvl}, we introduce a refinement trigger to encourage the model to revisit and correct its visual understanding. Specifically, we adopt a hybrid rollout strategy: for each group of $N$ samples, we randomly select $K$ samples and append a refinement trigger $g$ at the end of $C_{\mathrm{neg}}$. The trigger is sampled from a predefined trigger pool, with prompts such as \textit{``Wait, I may have misinterpreted the image.''} The remaining $N-K$ samples are kept unchanged. \textcolor{revisionblue}{As GRPO-style RL algorithms derive their learning signal from relative rewards within the same group of  rollouts, covering both behavior patterns, being prompted to re-examine the context and continuing  naturally, under the same erroneous context enables the model to better learn their relative advantage and produces  more stable reward differences. In our setup, we set $K = \frac{N}{2}$. In contrast, when $K$ is close to $0$ or $N$, the behaviors within a group become more homogeneous, weakening the relative differences available for credit assignment.}

This intra-group random injection encourages the model to learn a self-refinement behavior without supervised chain-of-thought data. Since triggers are only applied to samples with corrupted visual captions, the model avoids unnecessary reflection when no perceptual error is present. Through group-relative optimization, this mechanism encourages the model to internalize visual-space reasoning and correction capabilities.

\subsection{Policy Optimization and Reward Design}

In our training framework, the deceptive visual context is not directly used as a user prompt. Instead, it is inserted as an initial assistant prefix, which conditions the subsequent generation process and encourages the model to self-correct during reasoning. The full input sequence can be written as:
\begin{center}
\colorbox{cyan!20}{\textit{user:[image],[question]}}
\colorbox{lime!20}{\textit{assistant:<think>[deceptive context],[trigger]}}
\end{center}
where $\textit{<think> }$ is the predefined format token which indicates the beginning of the reasoning phase. During reinforcement learning, this assistant prefix is only used during rollout to influence the generation trajectory. It is not treated as an action in the policy optimization objective. Including the prefix in the action space would introduce off-policy bias and violate the standard policy gradient formulation, since the prefix is externally constructed rather than sampled from the policy distribution.

We adopt GRPO as the optimization algorithm. The reward function is defined as:
\begin{equation}
\label{eq:reward}
R_i = \alpha r_{\mathrm{format}} + \beta r_{\mathrm{acc}} + \gamma r_{\mathrm{cor}}\times r_{\mathrm{format}}\rev{\times r_{\mathrm{acc}}},
\end{equation}
where $r_{\mathrm{format}}$ enforces structured output in the form \textit{<think>} $\cdots$ \textit{</think> <answer>} $\cdots$ \textit{</answer>}, and $r_{\mathrm{acc}}$ is a binary reward indicating whether the final answer matches the ground truth. The correction reward $r_{\mathrm{cor}}$ evaluates whether the model correctly identifies and corrects visual misperceptions introduced in $C_{\mathrm{neg}}$. It is defined as:
\begin{equation}
r_{\mathrm{cor}}(X) =
\begin{cases}
0, & X \in \mathcal{D}_{\mathrm{0}}, \\
\frac{1}{|M|} \sum\limits_{(d_{\mathrm{before}}, d_{\mathrm{after}})\in M}
\mathbb{I}\Big(\Theta(X, d_{\mathrm{before}}, d_{\mathrm{after}})=1\Big),
& X \in \mathcal{D}_{\mathrm{aug}},
\end{cases}
\end{equation}
where $\Theta$ is a judge model that verifies whether the output explicitly rejects \rev{the modified erroneous description $d_{\mathrm{after}}$ and recovers the original visual evidence $d_{\mathrm{before}}$}. In Equation~\ref{eq:reward}, the correction reward is applied only when the answer is correct, preventing reward hacking on subjective or hard-to-verify responses.

%% file: sections/experiment.tex
\section{Experiments}

\textbf{Dataset} We train our model on two multimodal reasoning datasets, ViRL-39K~\citep{wangvl} and MMK12~\citep{meng2025mmeurekaexploringfrontiersmultimodal}. ViRL-39K emphasizes visual information and logic-centric reasoning tasks, while MMK12 focuses on mathematical and compositional reasoning. \rev{We sample approximately 18K valid examples from these datasets} \textcolor{revisionblue}{to form the original dataset $\mathcal{D}_0$, with the detailed sampling procedure provided in Appendix~\ref{appendix:data_construction}.} \textcolor{black} After applying the data construction pipeline (Sec.~\ref{sec:data_construction}), \rev{we retain the original samples and append their augmented versions}, with a ratio of $|\mathcal{D}_{\mathrm{aug}}| : |\mathcal{D}_{\mathrm{0}}| \approx 1:2$. We evaluate our method on both reasoning-intensive and vision-centric benchmarks. The former includes MathVista~\citep{lumathvista}, MathVerse~\citep{zhang2024mathversedoesmultimodalllm}, MathVision~\citep{wang2024measuringmultimodalmathematicalreasoning}, and WeMath~\citep{qiao-etal-2025-math}, while the latter includes HallusionBench~\citep{guan2024hallusionbench} (denote as ``HalluBench''), and FREAK~\citep{yin2026freak}. \rev{We also evaluate general multimodal reasoning on MMMU-Pro~\citep{yue-etal-2025-mmmu}.} For MMMU-Pro, we report the accuracy on the \textit{10-option} subset and \textit{vision} subset.

\textbf{Baselines} We compare against a comprehensive set of recent multimodal RL baselines, including Vision-R1~\citep{huang2026visionr}, VL-Rethinker~\citep{wangvl}, SophiaVL-R1~\citep{fan2026sophiavlr}, MM-Eureka~\citep{meng2025mmeurekaexploringfrontiersmultimodal}, R1-OneVision~\citep{yang2025r1}, Perception-R1~\citep{xiao2026perceptionr}, and PAPO$_{G}$~\citep{wang2026perceptionaware}. All methods are built upon the same backbone model, Qwen2.5-VL-7B~\citep{bai2025qwen25vltechnicalreport}, and trained using GRPO-style algorithm to ensure fairness.

\textbf{Implementation Details} For fair comparison, we adopt Qwen2.5-VL-7B as the backbone model. All models are trained for 2 epochs with a learning rate of $1 \times 10^{-6}$. The rollout batch size is set to 128, and the maximum generation length is 2048 tokens. For the reward function defined in Eq.~\ref{eq:reward}, we set the coefficients as $\alpha = 0.4$, $\beta = 1.0, \gamma = 0.2$. The expert model used for data construction is GPT-5.4. The intra-group triggers are defined as Figure~\ref{fig:prompt_triggers}. For more stable computation of accuracy reward, we use Qwen3-VL-32B as the judge model. During evaluation, we set the decoding temperature to 0.0 for all experiments.

\input{tables/main_table}

\subsection{Main Results}
As shown in Table~\ref{tab:main_results}, we present the main evaluation results of MIRA. Under the same parameter scale, our method achieves leading performance across multiple multimodal benchmarks, including MathVista (75.2), MathVision (29.0), WeMath (70.7), as well as non-mathematical reasoning benchmarks such as MMMU-Pro (41.0) and HallusionBench (69.8). On math tasks, MIRA achieves the best average performance compared to all baselines. For vision-centric benchmarks, whose task distributions are partially out-of-domain relative to the training data, our method still yields consistent performance improvements, demonstrating strong generalization ability. Overall, under the same model scale, MIRA exhibits clear advantages in both reasoning and visual understanding. These results highlight the importance of jointly improving reasoning capability and visual robustness in multimodal models. Detailed comparison between MIRA and vanilla GRPO is provided in Sec.~\ref{sec:ablation}.

\subsection{\rev{Generalization across Backbones}}
\label{sec:backbones}
\input{tables/backbone_results}
\rev{We additionally evaluate MIRA on Qwen2-VL-7B and Qwen3-VL-4B/8B. The first two use the original data and protocol; Qwen3-VL-8B uses difficulty-filtered ViRL-39K and MMK12, with the same data and training configuration for MIRA and GRPO. Table~\ref{tab:backbone_results} shows gains in 11 of 12 comparisons and a tie on Qwen3-VL-8B/MathVerse, supporting transfer across Qwen-VL generations and model sizes.}

\subsection{Ablation and Sensitivity Studies}
\label{sec:ablation}

\input{tables/ablation_data_reward}
\paragraph{Data Sensitivity} We first conduct a sensitivity analysis on the construction of augmented data. Specifically, we employ GPT-5.4, GPT-5.4-mini, and GPT-5.1-mini as expert models to rewrite the visual descriptions. As shown in the upper part of Table~\ref{tab:ablation_data}, using different expert models for description rewriting leads to varying degrees of improvement in MIRA's reasoning performance. In particular, stronger expert models yield better overall results, which may be attributed to their superior perception of the original images and their ability to produce higher-quality modified descriptions. In addition, we investigate model performance under different proportions of augmented data. The lower part of Table~\ref{tab:ablation_data} presents the evaluation results across varying data ratios; details of the dataset construction are provided in Appendix~\ref{appendix:more_experiment_results}. Among the selected ratios, the best performance is achieved when the proportion of augmented data is set to 50\%. In contrast, training without augmentation (i.e., vanilla GRPO) brings only limited improvements. Notably, when the proportion of augmented data becomes too high, performance degrades. We attribute this to the reduced diversity of training signals: although the model becomes better at handling misleading contexts, its intrinsic reasoning ability is not sufficiently improved. It is worth noting that reflection triggers are only introduced in the augmented data.

\paragraph{Reward and Intra-Group Trigger.}
\input{tables/ablation_triggers}
We further conduct a comprehensive ablation study to analyze the effects of both intra-group triggers and reward design on model performance. Table~\ref{tab:ablation_triggers} reports results under different configurations. We consider the following settings: (i) \textit{MIRA}, the full model with both correction reward and $K$ triggers in $N$ rollouts; (ii) \textit{w/o correction reward}, where the correction reward is removed while keeping the original trigger design; (iii) \textit{+ full trigger w/o correction reward}, where the correction reward is removed and triggers are applied to all samples within each group; (iv) \textit{w/o correction reward \& w/o trigger}, where both the correction reward and triggers are removed; and (v) \textit{GRPO}, which further removes the augmented context on top of (iv), reducing the method to standard GRPO training.  The results show that removing the correction reward already leads to noticeable performance degradation, indicating its role in encouraging effective error correction. As triggers are further simplified or removed, performance drops significantly, eventually approaching the level of vanilla GRPO. Without triggers, the model tends to ignore the injected deceptive context and instead rely on shortcut patterns to directly predict correct answers. Since reinforcement learning optimizes only the final answer reward, these shortcut-based solutions still receive high rewards. Consequently, such rollouts dominate the training signal, encouraging the model to disregard contextual information. This weakens its reliance on visual context and ultimately degrades its step-by-step reasoning ability.

\paragraph{MIRA Injection Strategy.}

\input{tables/inject_table}
The motivation of MIRA is to improve the model's reasoning robustness against misleading contextual information. To this end, we disguise incorrect context information as if it were generated by the model itself. To investigate its effect, in this ablation study we inject the perturbed context into the user prompt (represent as ``User'') in the following format:
\textit{"[image]. The content shown in the image above is: [deceptive contex]. The image description might contain errors. Please answer the question strictly based on the visual content: [question]"}. The correction reward is removed in this experiment. The comparison of training accuracy on the MathVista subset between our method (represent as ``Assistant'') and the variant without this injection strategy (``User'') is shown in Figure~\ref{fig:ablation_injection} and Table~\ref{tab:ablation_injection}. It can be observed that directly injecting deceptive context into the user prompt yields limited gains, with only marginal improvements compared to injecting it into the assistant prompt. Under this setting, the model fails to effectively enhance its perceptual and reasoning robustness from the erroneous context, and its performance on MathVista is nearly on par with vanilla GRPO training. Notably, the user-prompt injection method does not employ triggers.

%% file: tables/main_table.tex
\begin{table*}[t]
    \caption{Evaluation results of different reasoning methods. All results are reproduced using our evaluation scripts. Detailed evaluation settings are provided in Appendix~\ref{appendix:implementation_details}. Results of Vision-R1 on MathVision and WeMath are omitted due to potential training data contamination. Best (green) and second-best (blue) results among open-source multimodal reasoning methods are highlighted. Averages are computed over available benchmarks, excluding missing entries.}
    \label{tab:main_results}
    
    \centering
    \resizebox{1.0\textwidth}{!}
    {
        \renewcommand{\arraystretch}{1.5}
        \begin{tabular}{l c c c c c c c c}
            \toprule
            \multirow{2}{*}{} &  \multicolumn{4}{c}{\textbf{Math Benchmarks}} & \multicolumn{3}{c}{ \rev{\textbf{General / Visual Benchmarks}}} & \multirow{2}{*}{\textbf{Avg.}} \\
            \cmidrule(r){2-5} \cmidrule(l){6-8}
            & \textbf{MathVista} & \textbf{MathVerse} & \textbf{MathVision} & \textbf{WeMath} & \textbf{MMMUpro} & \textbf{HalluBench} & \textbf{FREAK} & \\
            \midrule
            \multicolumn{9}{l}{\textit{General-purpose Models}} \\
            \midrule
            GPT-4o & 62.6 & 50.4 & 31.6 & 67.6 & 53.2 & 73.5 & 33.4 & 53.2 \\
            InternVL2.5-8B & 59.6 & 34.8 & 19.2 & 56.3 & 33.2 & 67.5 & 29.7 & 42.9 \\
            InternVL3-8B & 71.6 & 47.7 & 29.5 & 62.9 & 41.9 & 67.5 & 29.7 & 50.1 \\
            Qwen2.5-VL-7B & 68.2 & 46.7 & 24.3 & 61.4 & 38.5 & 65.7 & 30.0 & 47.8 \\
            Qwen2.5-VL-32B & 76.1 & 55.6 & 36.8 & 70.9 & 41.2 & 72.3 & 26.7 & 54.2 \\
            \midrule
            \multicolumn{9}{l}{\textit{Multimodal Reasoning Methods}} \\
            \midrule
            
R1-Onevision-7B & 60.7 & 40.9 & 20.5 & 57.9 & 33.7 & 67.0 & 22.7 & 43.3 \\
Vision-R1-7B & 73.2 & 46.7 & - & - & 36.1 & 67.9 & 24.1 & 49.6 \\
MM-Eureka-7B & 71.2 & 50.6 & \second{\rev{28.7}} & 65.5 & 39.1 & 68.6 & 27.8 & 50.2 \\
PAPO$_{G}$-7B & 73.6 & 51.2 & \rev{27.7} & 66.3 & \best{41.1} & \second{68.7} & \best{31.1} & \second{51.4} \\
Perception-R1-7B & 71.9 & 50.9 & 27.5 & \best{70.7} & 40.7 & 67.5 & 24.8 & 50.6 \\
SophiaVL-R1-7B & 69.7 & 46.9 & 26.5 & 62.6 & 39.8 & \second{68.7} & 29.0 & 49.0 \\
VL-Rethinker-7B & \second{73.8} & \best{53.9} & 27.3 & 65.1 & 40.5 & 67.4 & 28.8 & 51.0 \\
\midrule
\textbf{Ours} & \best{75.2} & \second{53.8} & \best{29.0} & \best{70.7} & \second{41.0} & \best{69.8} & \second{30.6} & \best{52.9} \\
\rowcolor[gray]{0.9}$\Delta$ vs Qwen2.5-VL-7B & {\color[HTML]{008000} +7.0} & {\color[HTML]{008000} +7.1} & {\color[HTML]{008000} +4.7} & {\color[HTML]{008000} +9.3} & {\color[HTML]{008000} +2.5} & {\color[HTML]{008000} +4.1} & {\color[HTML]{008000} +0.4} & {\color[HTML]{008000} +5.0} \\
            \bottomrule
        \end{tabular}
    }
\end{table*}

%% file: tables/backbone_results.tex
\begin{table}[t]
\centering
\begin{revision}
\caption{\rev{Results on additional backbones (\%). Each MIRA/GRPO pair uses the same source data and training configuration.}}
\label{tab:backbone_results}
\papertablestyle
\begin{tabular*}{\linewidth}{@{\extracolsep{\fill}}llcccc@{}}
\toprule
\textbf{Backbone} & \textbf{Method} & \textbf{MathVista} & \textbf{MathVerse} & \textbf{MathVision} & \textbf{WeMath}\\
\midrule
\multirow{2}{*}{Qwen2-VL-7B} & GRPO & 62.0 & 35.6 & 18.9 & 48.1\\
 & \textbf{MIRA} & \textbf{63.5} & \textbf{36.5} & \textbf{19.1} & \textbf{49.5}\\
\midrule
\multirow{2}{*}{Qwen3-VL-4B} & GRPO & 77.1 & 59.6 & 45.4 & 75.6\\
 & \textbf{MIRA} & \textbf{77.8} & \textbf{61.6} & \textbf{47.0} & \textbf{78.7}\\
\midrule
\multirow{2}{*}{Qwen3-VL-8B} & GRPO & 78.0 & \textbf{66.1} & 51.1 & 79.1\\
 & \textbf{MIRA} & \textbf{79.9} & \textbf{66.1} & \textbf{52.9} & \textbf{81.2}\\
\bottomrule
\end{tabular*}
\end{revision}
\end{table}

%% file: tables/ablation_data_reward.tex
\begin{table}[t]
\centering
\caption{Comparison across benchmarks with different expert models and data ratio settings. Data ratio sensitivity experiment is conducted with 14K dataset. Refer Appendix~\ref{appendix:more_experiment_results} for details.}
\label{tab:ablation_data}
\papertablestyle
\begin{tabular*}{\linewidth}{@{\extracolsep{\fill}}lccccc@{}}
\toprule
\textbf{Configuration} & \textbf{MathVista} & \textbf{MathVerse} & \textbf{MathVision} & \textbf{WeMath} & \textbf{Avg.} \\
\midrule

\multicolumn{6}{c}{\textit{Expert Model Ablation (Expert Type)}} \\
\midrule
GPT 5.4      & \textbf{75.2} & \textbf{53.8} & 29.0 & \textbf{70.7} & \textbf{57.18} \\
GPT 5.4-mini & 74.3 & 52.5 & \textbf{29.7} & 69.6 & 56.53 \\
GPT 5.1-mini & 74.2 & 53.1 & 29.0 & 69.0 & 56.33 \\

\midrule
\multicolumn{6}{c}{\textit{Data Ratio Sensitivity (Augmentation Ratio)}} \\
\midrule
0 (GRPO$^{*}$) & 71.5 & 51.0 & 28.4 & 68.0 & 54.7 \\
0.3      & 72.5 & 52.2 & 28.5 & 67.9 & 55.3 \\
\textbf{0.5}      & \textbf{73.5} & \textbf{53.2} & 28.2 & \textbf{68.1} & \textbf{55.8} \\
0.7      & 71.5 & 51.3 & \textbf{28.9} & 68.0 & 54.9 \\

\bottomrule
\end{tabular*}
\end{table}

%% file: tables/ablation_triggers.tex
\begin{table}[t]
\centering
\caption{Ablation study on reward design and trigger mechanisms.}
\label{tab:ablation_triggers}

\papertablestyle

\begin{tabular*}{\linewidth}{@{\extracolsep{\fill}}lccccc@{}}
\toprule
\textbf{Setting} & \textbf{MathVista} & \textbf{MathVerse} & \textbf{WeMath} & \textbf{MathVision} & \textbf{Avg.} \\
\midrule
\textbf{MIRA (full)} & \textbf{75.2} & \textbf{53.8} & \textbf{70.7} & 29.0 & \textbf{57.2} \\
\midrule
w/o correction reward & 73.8 & 53.2 & \textbf{70.7} & 29.0 & 56.7 \\
\quad + full trigger & 73.6 & 52.6 & 68.8 & 28.4 & 55.8 \\
\quad - remove trigger & 70.9 & 51.8 & 68.8 & 29.0 & 55.1 \\
\quad - remove context (GRPO) & 72.9 & 51.5 & 68.8 & \textbf{29.3} & 55.6 \\
\bottomrule
\end{tabular*}

\end{table}

%% file: tables/inject_table.tex
\begin{table}[t]
\centering
\label{tab:user_vs_assistant}

\begin{minipage}[t]{0.48\linewidth}
\centering
\papertablestyle
\renewcommand{\arraystretch}{1.47} %

\captionof{table}{(a) Quantitative comparison of different context injection strategy. Training dataset is collected using GPT-5.4.}
\label{tab:ablation_injection}

\begin{tabular*}{\linewidth}{@{\extracolsep{\fill}}lcc@{}}
\toprule
\textbf{Benchmark} & \textbf{Assistant (Ours)} & \textbf{User} \\
\midrule
MathVista   & \textbf{73.8} & 70.7 \\
MathVerse   & \textbf{53.2} & 52.9 \\
WeMath      & \textbf{70.7} & 69.4 \\
MathVision  & \textbf{29.0} & 28.8 \\
MMMU        & \textbf{40.6} & 39.7 \\
\midrule
\textbf{Average} & \textbf{53.5} & 52.3 \\
\bottomrule
\end{tabular*}

\end{minipage}
\hfill
\begin{minipage}[t]{0.48\linewidth}
\centering

\captionof{figure}{(b) Evaluation dynamics on sampled 224 items from MathVista.}
\label{fig:ablation_injection}

\includegraphics[width=\linewidth]{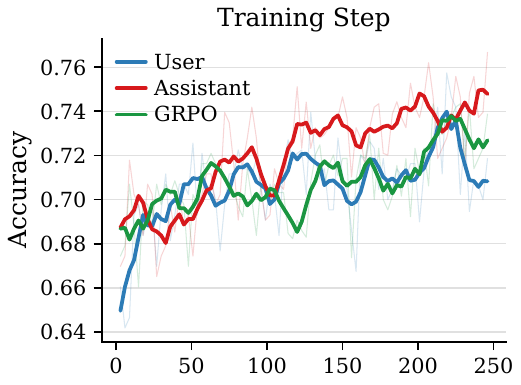}

\end{minipage}

\end{table}

%% file: sections/analysis.tex
\section{Analysis}
In this section, we provide a detailed analysis and case study of our method. We investigate why MIRA improves model performance compared to vanilla RL algorithms, and examine how the model's outputs evolve after being trained with MIRA.

\subsection{Method Analysis}
\begin{figure}[t]
\centering
\includegraphics[width=0.99\linewidth]{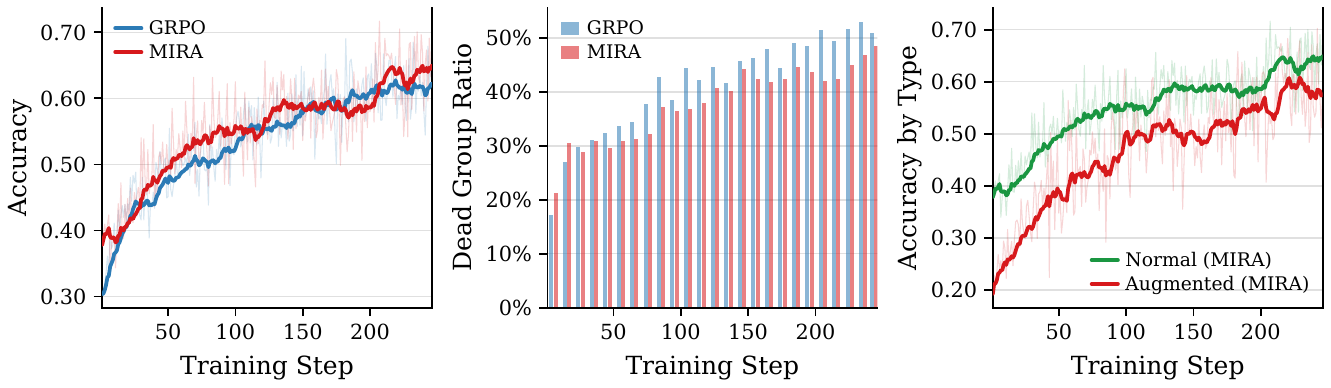}
\caption{Comparison between vanilla GRPO and MIRA. Left: training accuracy comparison; Middle: ratio of groups that the variance is 0; Right: accuracy comparison between normal samples and augmented samples. the reported result of MIRA is trained using GPT-5.4 as the expert model.}
\label{fig:analysis_dynamics}
\end{figure}
We analyze why MIRA achieves better performance than vanilla GRPO. From a methodological perspective, MIRA is designed to enhance the model’s reasoning capability by improving its robustness to hallucinated or misleading contextual information. Empirically, we observe that injecting incorrect context into originally easy samples significantly degrades model performance at the early stage of training. This enables such simple samples to provide non-trivial gradients and actively participate in model updates during MIRA training.  Furthermore, according to the design of Eq.~\ref{eq:reward}, for correctly answered samples, the correction reward further differentiates the rewards across different rollouts, thereby alleviating the issue where all responses within a group are correct and the gradient becomes zero. In contrast, under vanilla GRPO, these easy samples are typically answered correctly across all sampled responses, resulting in zero advantage for the entire group and thus zero gradients, which leads to slow or even stalled optimization. From the perspective of data augmentation, introducing augmented data increases the diversity of task types and implicitly raises the difficulty of the training data, mitigating the ineffectiveness of easy samples. As shown in the middle panel of Figure~\ref{fig:analysis_dynamics}, the number of sample groups with zero advantage variance is significantly reduced under MIRA compared to vanilla GRPO. This indicates that augmented data provides more informative training signals, validating the effectiveness of MIRA. 

When training with MIRA, the dataset can be divided into two subsets: \textit{normal data} and \textit{augmented data}. The left panel of Figure~\ref{fig:analysis_dynamics} shows the accuracy dynamics on normal data under both vanilla GRPO and MIRA. As training progresses, MIRA gradually outperforms GRPO on normal data. For the augmented data, the model progressively learns to handle misleading context, and its accuracy eventually becomes comparable to that on normal data, as illustrated in the right panel of Figure~\ref{fig:analysis_dynamics}.
\subsection{Case Study}
\begin{figure}[t]
\centering
\includegraphics[width=0.99\linewidth]{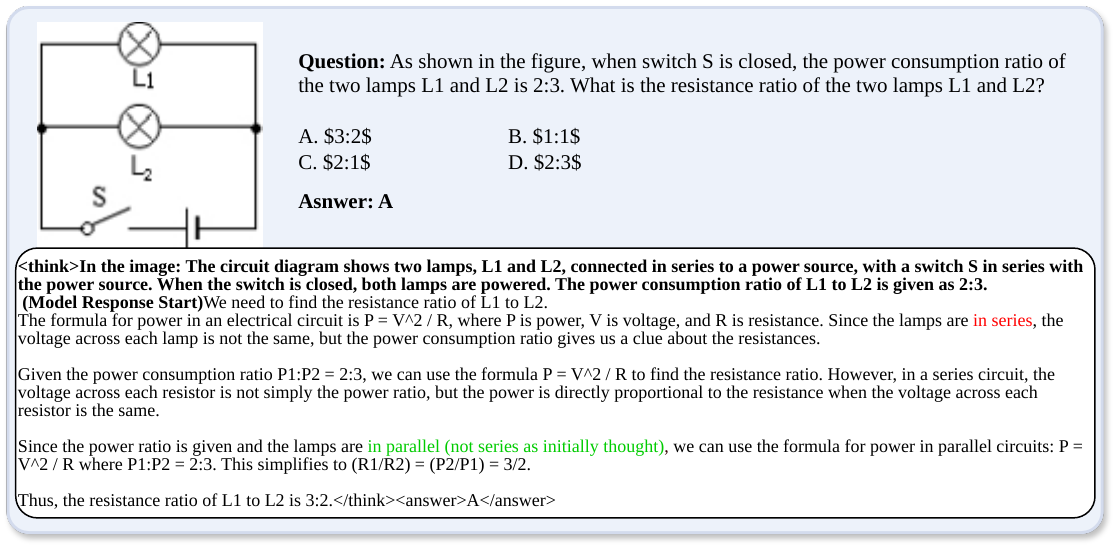}
\caption{Training case of MIRA. Notice that in this case no triggers are inserted. The model autonomously corrects both erroneous context and reasoning errors. Incorrect and correct parts are highlighted in red and green, respectively.}
\label{fig:case_study}
\end{figure}
Figure~\ref{fig:case_study} presents a training example along with the model’s output during training. In this sample, the input incorrectly describes the relationship between two resistors in the image, and this erroneous description is inserted at the beginning of the assistant’s response. No intra-group trigger is applied in this case. We observe that the model initially relies on the incorrect visual information for reasoning, but explicitly corrects these errors in the later stages of inference and arrives at the correct answer based on updated visual evidence. 

This case demonstrates that, even when the context contains corrupted or misleading visual information, the model does not directly rely on the erroneous textual descriptions associated with the visual input. Instead, it progressively re-evaluates and corrects its visual perception during reasoning, and ultimately leverages the correct visual evidence to produce its answer. This case suggests that, through MIRA training, the model exhibits a degree of visual robustness in long-form reasoning, enabling it to rectify incorrect visual cues even in later stages of the reasoning process, rather than merely relying on a superficial early-stage correction pattern.

\subsection{\rev{Quantitative Analysis of Reflection}}
\label{sec:reflection}
\input{tables/reflection_results}
\begin{revision}
We report reflection rate (RR), answer accuracy among reflections (SRR), and false correction rate (FCR) using a vision-capable LLM judge (Table~\ref{tab:reflection_results}). For $n$ inputs, let $z_i$, $a_i$, and $f_i$ indicate reflection, answer correctness, and false correction. The rates are $\mathrm{RR}=\sum_i z_i/n$, $\mathrm{SRR}=\sum_i z_i a_i/\sum_i z_i$, and $\mathrm{FCR}=\sum_i f_i/n$. Recall measures reflection on augmented inputs. Panel (a) samples augmented and normal groups separately, with SRR pooling both groups' reflections.

On augmented data, MIRA increases recall by 15.54 percentage points and SRR by 23.21 points, while improving both groups' accuracy. On MathVerse, RR increases from 8.9\% to 15.8\% and SRR from 57.8\% to 69.8\%; FCR also rises from 0.30\% to 1.07\%. On MathVista, accuracy improves from 72.9\% to 75.2\% despite lower SRR. These results complement the qualitative case, while showing that reflection can still introduce incorrect revisions.
\end{revision}

%% file: tables/reflection_results.tex
\begin{table}[t]
\centering
\begin{revision}
\caption{\rev{Quantitative reflection results on controlled and real inputs (\%).}}
\label{tab:reflection_results}
\papertablestyle
\begin{tabular*}{\linewidth}{@{\extracolsep{\fill}}llcccc@{}}
\toprule
\multicolumn{2}{l}{\textbf{(a) Controlled inputs}} & \textbf{Recall} & \textbf{SRR} & \textbf{Aug. acc.} & \textbf{Normal acc.}\\
\midrule
\multicolumn{2}{l}{Base} & 27.42 & 36.33 & 19.48 & 43.51\\
\multicolumn{2}{l}{\textbf{MIRA}} & \textbf{42.96} & \textbf{59.54} & \textbf{50.66} & \textbf{68.69}\\
\midrule
\textbf{(b) Benchmark} & \textbf{Method} & \textbf{Accuracy} & \textbf{RR} & \textbf{SRR} & \textbf{FCR}\\
\midrule
\multirow{2}{*}{MathVista} & GRPO & 72.9 & 4.1 & \textbf{85.4} & 0.30 (3/1000)\\
 & \textbf{MIRA} & \textbf{75.2} & 5.4 & 66.7 & \textbf{0.20} (2/1000)\\
\midrule
\multirow{2}{*}{MathVerse} & GRPO & 51.5 & 8.9 & 57.8 & \textbf{0.30} (12/3940)\\
 & \textbf{MIRA} & \textbf{53.8} & 15.8 & \textbf{69.8} & 1.07 (42/3940)\\
\bottomrule
\end{tabular*}
\end{revision}
\end{table}

%% file: sections/appendix.tex
\appendix

\section{Overview of the Appendix}

This appendix is organized as follows:
\textbf{Sec.~\ref{appendix:benchmarks}} discusses the benchmarks and baselines used for evaluation.
\textbf{Sec.~\ref{appendix:data_construction}} presents details of data construction.
\textbf{Sec.~\ref{appendix:implementation_details}} discusses implementation details, including training and evaluation procedures, and presents additional experimental results.
\rev{\textbf{Sec.~\ref{appendix:validity}} clarifies conditioning and optimization with external prefixes.}
\textbf{Sec.~\ref{appendix:prompt}} presents the prompts used during training and evaluation.

\section{Benchmarks and Baselines}
\label{appendix:benchmarks}
\subsection{Benchmarks}
\paragraph{MathVista} MathVista~\citep{lumathvista} is a unified benchmark for multimodal mathematical reasoning. We evaluate on its testmini split, which contains 1,000 problems across five subtasks: Textbook Question Answering, Visual Question Answering, Geometry Problem Solving, Math Word Problems, and Figure Question Answering. Compared to other math benchmarks, MathVista places greater emphasis on visual information extraction, and the logical difficulty of its problems is lower.

\paragraph{MathVerse} MathVerse~\citep{zhang2024mathversedoesmultimodalllm} is a benchmark designed to evaluate the reasoning abilities of multimodal large language models (MLLMs) under varying proportions of textual and visual information. The testmini split of MathVerse contains 3,940 problems, including multiple-choice and fill-in-the-blank questions. It is further divided into five subsets based on the balance between text and visual content: Text Dominant (TD), Text Lite (TL), Vision Intensive (VI), Vision Dominant (VD), and Vision Only (VO). MathVerse covers three subtasks: Plane Geometry (2,550 problems), Functions (795 problems), and Solid Geometry (595 problems).

\paragraph{MathVision} MathVision~\citep{wang2024measuringmultimodalmathematicalreasoning} consists of 3,040 high-quality mathematical problems with visual contexts sourced from real mathematical competitions. In terms of mathematical reasoning difficulty, it is more challenging than both MathVista and MathVerse. The dataset includes five difficulty levels and 16 subtasks. In our evaluation, we do not differentiate between difficulty levels or subtasks.

\paragraph{WeMath} WeMath~\citep{qiao-etal-2025-math} is a benchmark specifically designed to explore problem-solving principles beyond end-to-end performance. It covers 67 hierarchical knowledge concepts across five levels of knowledge granularity. All questions in WeMath are multiple-choice, and its testmini split contains 1,740 problems in total. The dataset emphasizes the understanding and application of existing mathematical knowledge as well as visual perception.

\paragraph{MMMU-Pro} MMMU-Pro~\citep{yue-etal-2025-mmmu} is a carefully filtered version of the multimodal benchmark MMMU. It improves upon MMMU in three aspects: (1) removing questions that can be answered solely by text-based models; (2) increasing the number of multiple-choice options, making random guessing more difficult; and (3) introducing a vision-only setting, where questions are embedded within images for evaluation. In our experiments, we use the Vision split and the 10-Options split of MMMU-Pro, each containing 1,730 questions. We compute accuracy on both splits separately and report the final MMMU-Pro overall result as the average accuracy across the two settings.

\paragraph{HallusionBench} HallusionBench~\citep{guan2024hallusionbench} is a widely used benchmark for evaluating multimodal hallucinations. It contains 346 images and 1,129 questions. The images are carefully edited by human experts to violate real-world or visual commonsense, thereby intentionally inducing hallucinations in multimodal models through conflicts with their internal textual knowledge.

\paragraph{FREAK} FREAK~\citep{yin2026freak} is a fine-grained hallucination evaluation benchmark designed for next-generation multimodal large language models. It includes 1,799 diverse images and corresponding questions. Compared to HallusionBench, FREAK constructs counterfactual images using an automated pipeline. These images are of higher resolution, focus on more complex real-life scenarios, and involve localized modifications to object attributes. Existing studies show that state-of-the-art multimodal models perform poorly on FREAK.

\subsection{Baselines}

\input{tables/baseline_comparison}
To ensure a fair comparison with the baselines, Table~\ref{tab:baselines} summarizes the basic information of the baselines considered in this paper, including their venue and publication year, as well as the scale and sources of their training data. According to the table, the data scale used by MIRA is comparable to or slightly smaller than that of the baselines, and all the data are sourced from existing open-source works, which helps ensure the fairness of the comparison to a certain extent.

\section{Data Construction}
\label{appendix:data_construction}
The construction of visual error data is a core component of our method. During training, we use a mixture of original samples and preprocessed visual error data, both derived from existing datasets. We first introduce the datasets used, followed by a detailed description of the data processing pipeline.

\begin{figure}[t]
\begin{center}
\includegraphics[width=0.7\linewidth]{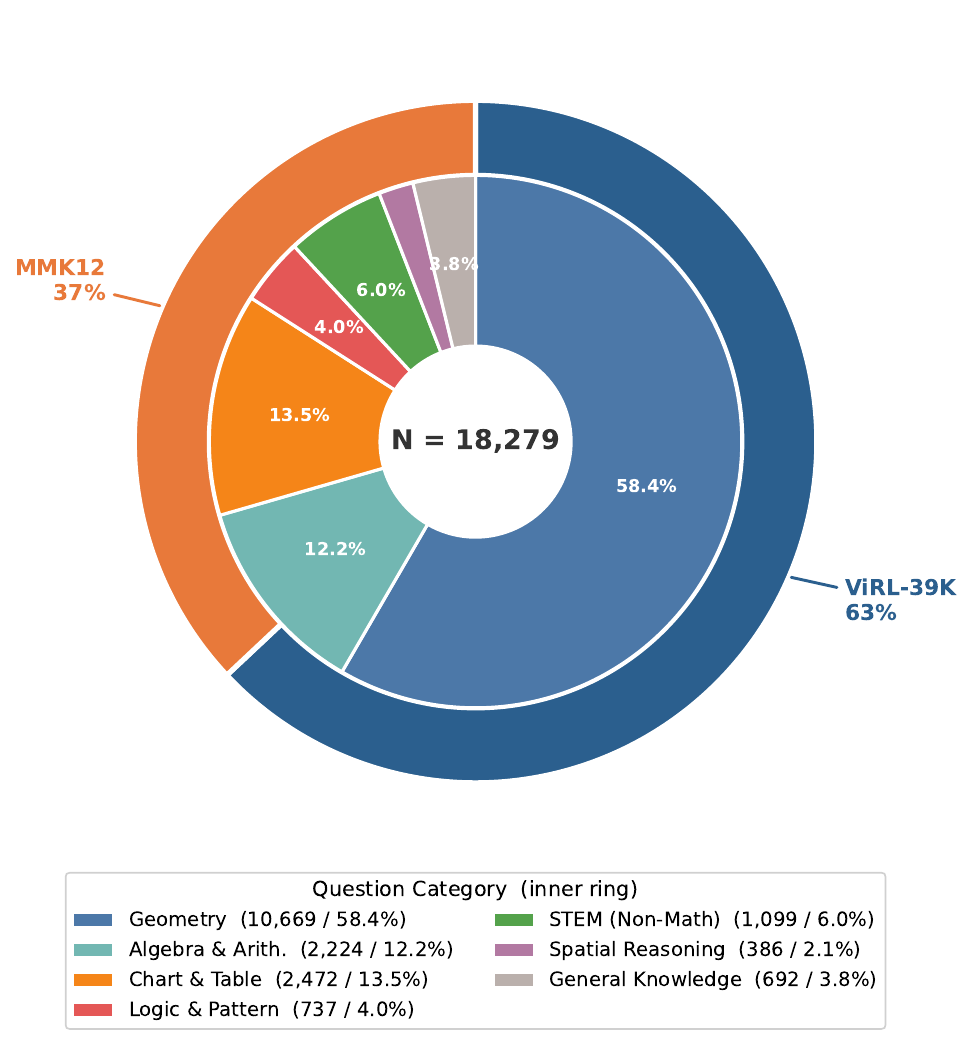}
\end{center}
\caption{The statistics of training data.}
\label{fig:data_statistics}
\end{figure}

\rev{We use ViRL-39K~\citep{wangvl} and MMK12~\citep{meng2025mmeurekaexploringfrontiersmultimodal}. We initially sample 20K instances, with 63\% from ViRL-39K and 37\% from MMK12, and remove instances with invalid images to obtain an approximately 18K-example source pool. Figure~\ref{fig:data_statistics} shows its composition.}

We then preprocess the data by introducing a sampling probability $p$. \rev{An augmented counterpart is constructed for each original sample with probability $p$}, and is otherwise left without an augmented counterpart. For samples selected for augmentation, we first use a policy model (Qwen2.5-VL-7B in this work) to generate an initial \rev{question-relevant visual context, which is not guaranteed to be fully correct}. Specifically, given an image and a question, the model is prompted to first perform visual understanding by extracting all information relevant to the question (i.e., generating a question-related caption), and then produce the final answer. We retain only the visual understanding component, denoted as $C_{\mathrm{pos}}$, and discard the answer.

\begin{figure}[h]
\begin{center}
\includegraphics[width=0.99\linewidth]{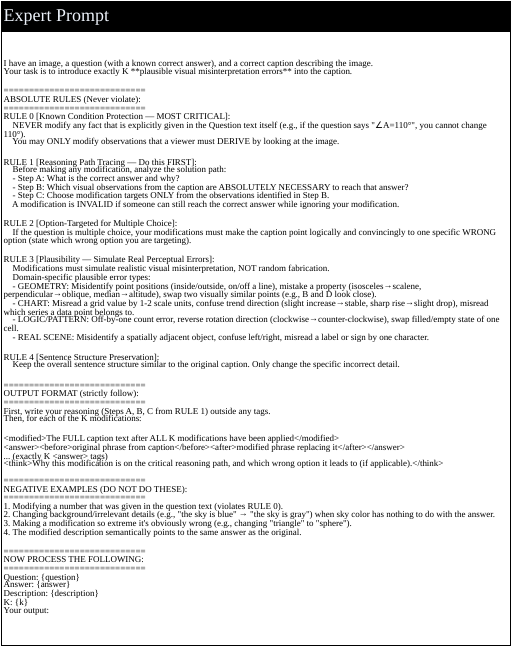}
\end{center}
\caption{The prompt we used for expert modification in data construction.}
\label{fig:prompt_expert_modification}
\end{figure}

Next, we employ an expert model to transform $C_{\mathrm{pos}}$ into an incorrect version $C_{\mathrm{neg}}$ using a carefully designed instruction prompt (Fig.~\ref{fig:prompt_expert_modification}). The prompt guides the model to introduce targeted errors into key visual information while performing self-verification to ensure the effectiveness of the modifications. The expert model is also required to provide structured explanations, including explicit annotations of the original and modified text. We extract these modifications using regular expressions to construct a modification set $M$, where each entry records detailed before-and-after changes.

Due to model limitations, not all differences between $C_{\mathrm{pos}}$ and $C_{\mathrm{neg}}$ are relevant to solving the task. For example, if the question concerns region A while the modification occurs in an irrelevant region B, the injected error does not affect the answer. To eliminate such ineffective perturbations, we further validate $C_{\mathrm{neg}}$. Specifically, we construct two inputs: (a) question $Q$ with correct context $C_{\mathrm{pos}}$, and (b) question $Q$ with incorrect context $C_{\mathrm{neg}}$. A sample is retained only if the verification model answers correctly in (a) and incorrectly in (b).

Importantly, we do not use the policy model as the verifier. Since the policy model is relatively weak during training, it may fail even when provided with correct context, leading to the removal of valid samples. As the purpose of this step is to identify truly misleading $C_{\mathrm{neg}}$, using a stronger model significantly improves filtering quality and data efficiency.

\rev{We retain the entire original pool $\mathcal{D}_0$ and append the filtered augmented counterparts $\mathcal{D}_{\mathrm{aug}}$. Thus $\mathcal{D}_{\mathrm{MIRA}}=\mathcal{D}_0\cup\mathcal{D}_{\mathrm{aug}}$, with an augmented-to-original instance ratio of approximately $1:2$. Augmented instances reuse source image--question pairs and do not add new questions. The separate 14K ratio study below uses a reconstructed pool and should not be confused with this main mixture.}

\rev{Although $C_{\mathrm{pos}}$ is not intentionally corrupted, it may contain policy perception errors. Answer-based filtering selects task-relevant perturbations but does not exhaustively validate the description. The correction reward evaluates only the explicitly recorded local errors; unannotated background errors are not directly scored by this term.}
\section{Implementation Details}
\label{appendix:implementation_details}
\subsection{Training Details}
\input{tables/hyperparam}
We adopt veRL as the reinforcement learning framework. During GRPO training, rule-based verification is first performed via regular expression matching; if this step fails, a judge model is used for further evaluation. In our implementation, we employ Qwen3-VL-32B as the judge model. Notably, both the policy model and the judge model are co-deployed on the same GPUs, \rev{without requiring additional dedicated GPU devices.} The detailed hyperparameter settings for reinforcement learning are provided in Table~\ref{tab:training_hyperparams}. Due to disk I/O bottlenecks during training, the entire training process takes approximately 16 hours.

\subsection{Evaluation Details}

We evaluate model performance using a custom-built evaluation pipeline. Across all benchmarks, we employ a large language model (GPT-4o) as the judge model. Specifically, we first extract the final answer from model outputs using regular expressions; when extraction fails or further verification is needed, the judge model is used to determine whether the predicted answer matches the ground truth. Due to the inherent stochasticity of API-based outputs, this evaluation process introduces minor randomness. This evaluation variability is distinct from variation across independent training runs.

\begin{figure}[h]
\begin{center}
\includegraphics[width=0.99\linewidth]{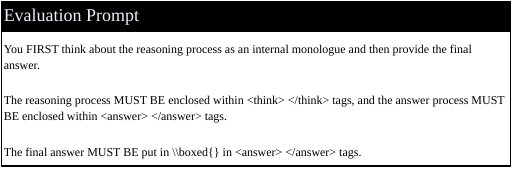}
\end{center}
\caption{The system prompt we used for evaluation.}
\label{fig:prompt_evaluation}
\end{figure}

To ensure fairness in evaluation, the system prompts used for different benchmarks are not derived from those used during training. Instead, we adopt a unified system prompt (shown in Fig.~\ref{fig:prompt_evaluation}) across all selected benchmarks. As this prompt is not tailored to any specific baseline, it provides a fair and consistent evaluation setting.

\subsection{More Experiment Results}
\label{appendix:more_experiment_results}
In Table~\ref{tab:ablation_data}, the ablation study on expert models uses the data described in Appendix~\ref{appendix:implementation_details}. However, for the data ratio sensitivity analysis, we reconstruct the dataset based on the above data. Specifically, we first merge the augmented data generated by three expert models, namely GPT-5.4, GPT-5.4-mini, and GPT-5.1-mini. For each sample, we select its augmented version following the priority order GPT-5.4 → GPT-5.4-mini → GPT-5.1-mini. If a sample is a normal (non-augmented) instance in all three datasets, it is discarded. This process results in a fully augmented dataset of 14K samples (i.e., 100\% augmented). We then perform random degradation by converting a portion of the augmented samples back to normal samples, thereby constructing datasets with augmented data ratios of 0.7, 0.5, 0.3, and 0. So, in this table, GRPO$^{*}$ denotes training on the 14K dataset, which differs from the data used for vanilla GRPO in the other tables. \rev{The verified four-benchmark expert comparison is reported in Table~\ref{tab:ablation_data}.}

\Needspace{6\baselineskip}
\section{\rev{Optimization and Implementation Clarification}}
\label{appendix:validity}
\begin{revision}
This section clarifies external-context conditioning rather than claiming a new theoretical result. A corrupted description is generated offline and fixed during RL training. A trigger is also supplied externally, with a selection rule independent of policy parameters. Neither is a policy-sampled action.

Let $x=(I,Q)$, $c=C_{\mathrm{neg}}$, and $g_i$ be the supplied trigger for rollout $i$ (empty for untriggered rollouts). The state is $s_i=(x,c,g_i)$, and the sampled continuation follows
\begin{equation}
\pi_\theta(y_i\mid s_i)=\prod_{t=1}^{T_i}\pi_\theta(y_{i,t}\mid s_i,y_{i,<t}).
\end{equation}
The policy and behavior-policy likelihoods must condition on the same supplied prefix. At a generated token, the corresponding ratio is
\begin{equation}
\rho_{i,t}=\frac{\pi_\theta(y_{i,t}\mid s_i,y_{i,<t})}
{\pi_{\theta_{\mathrm{old}}}(y_{i,t}\mid s_i,y_{i,<t})}.
\end{equation}
The loss mask excludes the image/question prompt, injected description, and externally supplied trigger; generated continuation tokens contribute to the clipped GRPO objective. Treating fixed prefix tokens as sampled actions would incorrectly attach a policy likelihood to externally provided text. Masking those tokens does not prevent the continuation loss from depending on the prefix through its contextual representations.

This observation concerns the likelihood accounting for the injected context. It does not establish that a clipped GRPO surrogate with group-normalized advantages is an unbiased estimator of an unclipped expected-reward gradient. It also does not eliminate the distribution gap between synthetic descriptions and naturally occurring perception errors. MIRA targets robustness to controlled misleading contexts; the unaugmented benchmark analysis assesses behavioral transfer separately.
\end{revision}

\section{\rev{Additional Behavioral Analyses}}
\subsection{\rev{Trigger Wording Sensitivity}}
\label{appendix:trigger_sensitivity}
\begin{revision}
We evaluate the trained Qwen2.5-VL-7B policy on 500 additional augmented examples with no trigger and with each of seven English triggers. The trigger pool also contains Chinese templates to match the training data's language distribution. Table~\ref{tab:wording_sensitivity} reports the mean and standard deviation across English phrasings; these are not independent training runs. Final-answer accuracy varies less across phrasings than reflection frequency. This experiment tests prompted behavior at evaluation time, whereas standard benchmark evaluation uses no trigger.
\begin{table}[ht]
\centering
\caption{\rev{Trigger wording sensitivity on 500 augmented examples (\%). Variability is across seven English phrasings.}}
\label{tab:wording_sensitivity}
\papertablestyle
\color{revisionblue}
\begin{tabular*}{\linewidth}{@{\extracolsep{\fill}}lcc@{}}
\toprule
\textbf{Metric} & \textbf{No trigger} & \textbf{Seven triggers (mean $\pm$ std.)}\\
\midrule
Reflection rate & 42.2 & $90.36\pm3.22$\\
SRR & 39.8 & $\boldsymbol{56.64\pm1.07}$\\
Accuracy & 42.2 & $\boldsymbol{54.73\pm1.18}$\\
\bottomrule
\end{tabular*}
\end{table}
\end{revision}

\subsection{\rev{Reflection Metrics and Interpretation}}
\begin{revision}
\paragraph{Sampling and generation.}
The evaluator reads two groups from a parquet file: type~1 supplies an erroneous description as an unfinished assistant prefix, while type~0 ends with a user prompt and contains no injected prefix. It samples each group separately using a fixed random seed, stores a sample manifest, and evaluates every checkpoint on exactly the same inputs. The two groups are not assumed to be question-paired. The groups are deterministically shuffled together before generation. The supplied evaluator uses greedy decoding (temperature~0, top-$k=1$); the controlled comparison calls generation without an added reflection trigger.

\paragraph{Vision-capable LLM judge.}
For each response, the judge receives the image, question, reference answer, and model continuation. For augmented inputs it also receives the injected description and each wrong-claim/correct-evidence pair. It labels explicit reflection, the number of corrected modifications, whether all modifications are corrected, and final-answer correctness independently, with a supporting quote and explanation. Generic phrases such as ``let me check,'' routine step-by-step reasoning, or a correct answer without an expressed revision do not qualify as reflection. For normal inputs, reflection requires reconsideration of an earlier belief, calculation, or visual interpretation in the response itself.

\paragraph{Aggregation.}
Reflection recall is the fraction of valid augmented responses with reflection. The answer-based SRR pools all reflected responses from both controlled groups. Augmented and normal accuracies are computed separately. A stricter successful-correction label additionally requires explicit correction of every injected modification and a correct final answer; it is not the SRR column. Judge failures are excluded from valid-response denominators. On normal inputs, the strict injected-modification correction label is inapplicable and set to false. Reflection on normal inputs may be legitimate self-correction: counting it as a false positive for injected-error detection does not make it a false correction.

For the unaugmented benchmark analysis, FCR divides erroneous-revision counts by the full benchmark size. This is distinct from the controlled evaluator's detection false positives. Counts are retained in Table~\ref{tab:reflection_results} to make the normalization explicit. The conditional statistics do not identify a causal effect of reflection, since the models may reflect on different subsets.
\end{revision}

\subsection{\rev{User-Side Injection Prompt}}
\label{appendix:injection_prompt}
\rev{The user-side ablation uses: ``[image]. The content shown in the image above is: [deceptive context]. The image description might contain errors. Please answer the question strictly based on the visual content: [question].'' This setting omits reflection triggers and the correction reward.}

\Needspace{7\baselineskip}
\subsection{\rev{Failure Modes}}
\rev{The observed failure modes include continuing an incorrect reasoning path after revising a visual interpretation, and producing inconsistent perceptions before and after reflection on clean inputs. These two patterns distinguish recognizing a discrepancy from using the corrected information consistently. The rise in false corrections on MathVerse provides quantitative evidence of erroneous revisions; the successful trajectory in Figure~\ref{fig:case_study} should not be interpreted as a reliability estimate.}

\section{Used Prompts}
\label{appendix:prompt}
The following presents the prompts used throughout the entire process of MIRA.

\begin{figure}[h]
\begin{center}
\includegraphics[width=0.99\linewidth]{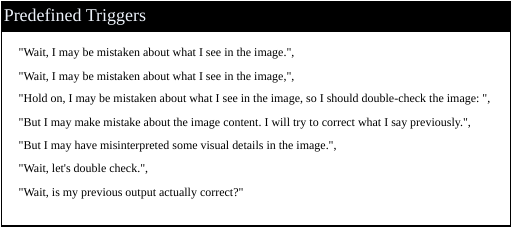}
\end{center}
\caption{The pre-defined triggers set of intra-group trigger.}
\label{fig:prompt_triggers}
\end{figure}

\begin{figure}[h]
\begin{center}
\includegraphics[width=0.99\linewidth]{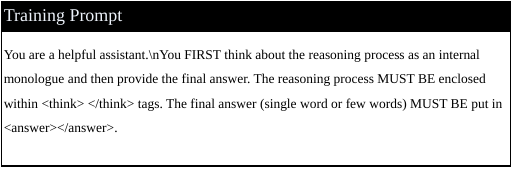}
\end{center}
\caption{The system prompt we used for training.}
\label{fig:prompt_training}
\end{figure}

%% file: tables/baseline_comparison.tex
\begin{table}[t]
\centering
\caption{Summary of baselines, including venue, data scale, and data source.}
\label{tab:baselines}
\papertablestyle
\begin{tabular*}{\linewidth}{@{\extracolsep{\fill}}lcc@{}}
\toprule
\textbf{Baseline} & \textbf{Venue} & \textbf{Data Scale} \\
\midrule
R1-Onevision~\citep{yang2025r1} & ICCV'2025  & R1-Onevision-155K  \\
Vision-R1~\citep{huang2026visionr} & ICLR'2026  & 200K SFT + 10K RL  \\
MM-Eureka~\citep{meng2025mmeurekaexploringfrontiersmultimodal} & - & MMK12 (15.6K)  \\
PAPO~\citep{wang2026perceptionaware} & ICLR'2026 & ViRL-39K \\
Perception-R1~\citep{xiao2026perceptionr} & ICLR'2026 & Geometry3K \\
SophiaVL-R1~\citep{fan2026sophiavlr} & ICLR'2026 & SophiaVL-130K \\
VL-Rethinker~\citep{wangvl} & NeurIPS'2025 & ViRL-39K \\
\midrule
\textbf{MIRA (Ours)} & - & ViRL+MMK12 (18K)  \\
\bottomrule
\end{tabular*}
\end{table}

%% file: tables/hyperparam.tex
\begin{table}[h]
\centering
\caption{Training hyperparameters we used for main results.}
\label{tab:training_hyperparams}
\papertablestyle
\begin{tabular*}{\linewidth}{@{\extracolsep{\fill}}lclc@{}}
\toprule
\multicolumn{2}{c}{\textbf{Training Configuration}} & \multicolumn{2}{c}{\textbf{RL Configuration}} \\
\cmidrule(r){1-2} \cmidrule(l){3-4} 
Training epochs             & 2         & Rollout number & 8      \\
Gradient accumulation steps & 4         & KL Coefficient & 0.004  \\ 
Effective batch size        & 128       & Rollout Engine & vLLM\\ 
Sequence length cutoff      & 5,000     & Temperature & 1.0\\
\cmidrule(l){3-4} 
Optimizer                   & AdamW     &   \multicolumn{2}{c}{\textbf{Trigger Configuration}} \\
\cmidrule(l){3-4} 
Learning rate               & $1 \times 10^{-6}$ &  \rev{Rollouts with trigger $K$} & \rev{4} \\
Max gradient norm           & 1.0       &         \rev{Rollouts per group $N$} & \rev{8}                    \\
Precision                   & bfloat16  &      \rev{Augmented:clean} & \rev{$\approx1:2$}                       \\
\bottomrule
\end{tabular*}
\end{table}